\documentclass[10pt, conference, compsoc]{IEEEtran}
\usepackage{graphicx}
\usepackage[tight,footnotesize]{subfigure}
\usepackage{booktabs}
\usepackage{textcomp}
\usepackage{algorithm,algorithmic}
\usepackage{amsmath}
\usepackage{wrapfig}
\usepackage{verbatim}
\usepackage{placeins}
\usepackage{cite}
\usepackage{etoolbox}
\makeatletter
\patchcmd{\@makecaption}
  {\scshape}
  {}
  {}
  {}
\makeatother

\newcommand{\CASE}[1]{\STATE \textbf{case} #1\textbf{:} \begin{ALC@g}}
\newcommand{\ENDCASE}{\end{ALC@g}}

\newcommand{\DEFAULT}{\STATE \textbf{default:} \begin{ALC@g}}
\newcommand{\ENDDEFAULT}{\end{ALC@g}}
\newcommand{\DEFAULTLINE}[1]{\STATE \textbf{default:} }
\usepackage{listings}

\usepackage[tight,footnotesize]{subfigure}
\usepackage[table]{xcolor}
\usepackage{array,hhline}
\usepackage{pgfplots}
\usepackage{anyfontsize}
\makeatletter
\newcommand*{\ccol}[1]{%
  \ifdim#1pt<.5pt\relax\else\color{white}\fi
  \edef\x{\noexpand\cellcolor[gray]{\strip@pt\dimexpr1pt-#1pt}}\x
  #1%
}
\newlength{\cellwidth}
\makeatother
\usepackage[english]{babel}
\usepackage{fancyhdr}
\usepackage{lastpage}
 
\pgfplotsset{compat=1.14} 

\begin{document}

\title{Epileptic seizure classification using statistical sampling and a novel feature selection algorithm}

\author{\IEEEauthorblockN{ Md Mursalin$^1$, Syed Mohammed Shamsul Islam$^{1}$, Md Kislu Noman$^2$, Adel Ali Al-Jumaily$^{1,3}$}
\IEEEauthorblockA{$^{1}$Edith Cowan University, Australia\\
$^2$Pabna University of Science and Technology, Bangladesh\\
$^3$University of Technology Sydney, Australia\\
 }
}
\maketitle

\begin{abstract}
Epilepsy is one of the most common neuronal disorders that can be identified by interpretation of the electroencephalogram (EEG) signal. Usually, the length of an EEG signal is quite long which is challenging to interpret manually. In this work, we propose an automated epileptic seizure detection method by applying a two-step minimization technique: first, we reduce the data points using a statistical sampling technique and then, we minimize the number of features using our novel feature selection algorithm. We then apply different machine learning algorithms for evaluating the performance of the proposed feature selection algorithm. The experimental results outperform some of the state-of-the-art methods for seizure detection using the reduced data points and the least number of features.

\end{abstract}
\IEEEpeerreviewmaketitle
\section{Introduction}
\label{introduction}
Epilepsy can be identified by recurrent seizure activity \cite{orosco2016patient}\cite{CHAIBI2013927}. An epileptic seizure is an abnormal neuronal activity in the brain \cite{li2016sequential,xiang2017local}. Such seizures may cause a severe effect on the cognitive system of humans \cite{sanei2013eeg, parvez2014epileptic, kolekar2015nonlinear}. The abnormal activity can be measured by monitoring electrical impulses on the surface area of the cortex. Electroencephalography (EEG) is one of the powerful clinical tools for epilepsy detection \cite{ngugi2011incidence, subasi2006comparison}. Generally, the length of an EEG signal is quite long and requires much time to measure \cite{pippa2016improving}. Therefore, an efficient computer-based method can provide a feasible solution \cite{birjandtalab2017automated}.

\begin{figure*}[htb] 
\centering
\includegraphics[height=1.5cm,width=1\linewidth]{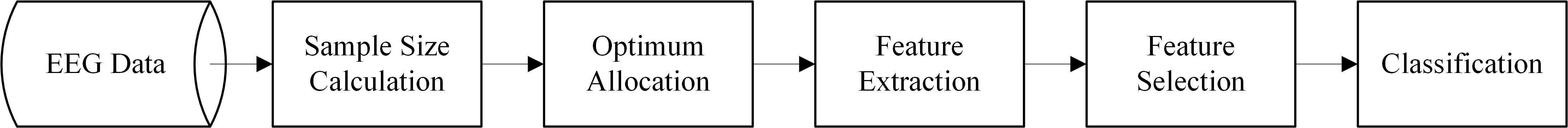}
\caption{\label{f1}The block diagram of the proposed method.}
\end{figure*}

The statistical sampling technique for biomedical signal processing is quite a new idea. This technique was first introduced in brain-computer interface (BCI) application by Siuly et al. \cite{siuly2015discriminating}. Firstly, they divided the whole signal into smaller groups named strata and calculated the sample size for each group named stratum. Next, they merged each stratum to construct the reduced signal data. From this reduced signal data, authors extracted different features and applied a least support vector machine and Naive Bayes classifier for classifying motor imagery tasks. A similar strategy was used by Siuly et al. \cite{siuly2015exploring} for detection of multi-category EEG signals. They applied three different conventional classification algorithms including SVM, k-NN, and multinominal logistic regression with a ridge estimator for classification. Kabir et al. \cite{kabir2016epileptic} also proposed a sampling-based technique for epileptic seizure detection from multi-class EEG signals. We are mostly inspired by their idea of applying statistical sampling technique to analyze the EEG signal. However, all the stated approaches only considered linear features that may not be able to extract the hidden subtle changes in the time series \cite{acharya2012automated}\cite{wang2016epileptic}. They did not apply any feature selection method to reduce the feature set and did not show the impact of changing different confidence levels for data reduction. 

In this paper, we investigate two key challenges for epilepsy detection: size of data, and the number of features. So, we develop our method in such a way that it can reduce both the size of data and the number of features. To reduce the data size, we apply a statistical sampling technique called optimum sample allocation technique (OA). For reducing the required features, we develop a feature selection algorithm. The contribution of this work can be summarized as follows:

\begin{enumerate}
  \item    Developing a novel feature selection method 
  \item Minimizing data using the statistical sampling technique
  \item Analysing the performance using different classification algorithms: Support Vector Machine (SVM), Random Forest (RF), Naïve Bayes (NB), K-Nearest Neighbor (KNN) and Logistic Model Trees (LMT)
\end{enumerate}
The rest of this paper is structured as follows. Section 2 describes the proposed method. Section 3 explains the experimental results, and Section 4 provides the concluding statements.

\section{Materials and methods}

In this work, we use a sampling technique for data reduction and develop a feature selection algorithm to select the least number of features. So, first, we estimate the sample size using a different confidence level. Next, we divide the signal into smaller segments known as strata and calculate the sample size for each stratum (singular of strata) using the optimum allocation technique. We then select the sample known as the optimum allocated sample from each stratum. After selecting sample data, we extract features from each stratum. We then apply our proposed feature selection algorithm and evaluate the performance using different classifiers. The block diagram of our proposed method is shown in Figure~\ref{f1}.

\subsection{Dataset}
\begin{figure} [htb]
\centering
\includegraphics[width=1\linewidth]{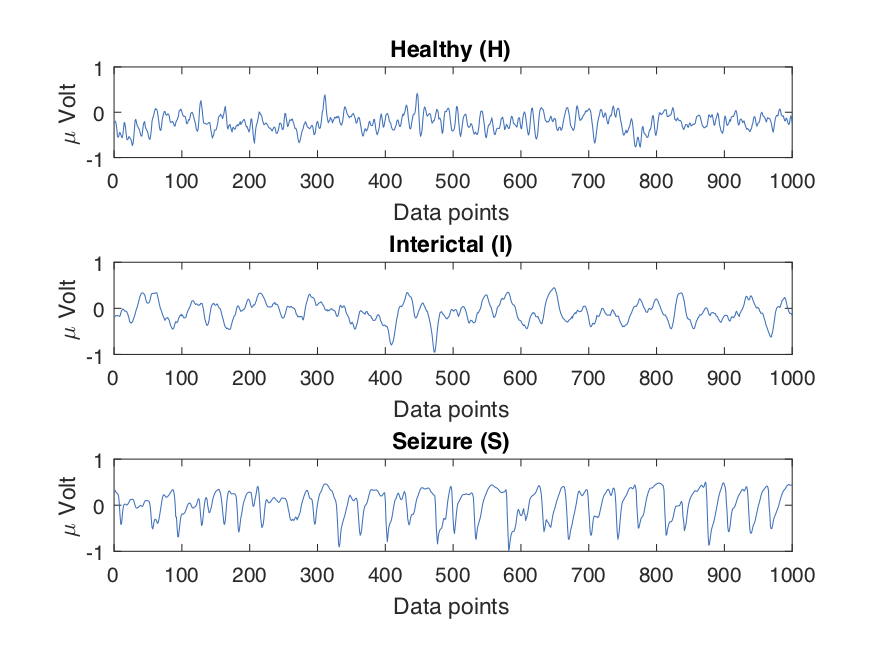}
\caption{\label{f2}Sample EEG signal data from three different groups named as healthy, interictal and seizure.}
\end{figure}

In this work, we use benchmark EEG data from University of Bonn \cite{andrzejak2001indications}. The entire database contains five datasets named as set A to E. Each set includes one hundred individual channels. The length of each channel is 23.6 second. The number of data points in each channel is 4097. Set A and B are obtained from the surface EEG signals of five healthy participants, where set A consists of the signal during open eyes and the set B consists of the signal during closed eyes. Apart from set A and B, all the other sets are collected from five patients who had epilepsy. Both set C and D are taken from seizure-free intervals. Set C is collected from the opposite hemisphere of the brain and set D is collected from the epileptogenic zone. Only set E includes the signal with seizure activity. For all the EEG data, an average common reference with a 128-channel amplifier system is used where the sampling rate is 173.61 Hz. The signals are preprocessed by 0.53-40 Hz band-pass filter. The noise due to the eye and muscle movements are eliminated by visual inspection. Sample EEG signals from three different groups denoted as healthy, interictal, and seizure are illustrated in Figure~\ref{f2}.

\subsection{Sample size calculation}
The first step for stratified sampling is to find the sample size. In this paper, the entire signal data is expressed by the population, and the representative data from the population is named as the sample. We calculate the sample size using Equation (1) and (2)  \cite{siuly2015discriminating}\cite{cochran2007sampling},
\begin{equation}
n=\frac{z^2\times p \times(1-p)}{e^2}
\end{equation}
where $n$ is the sample size, $z$ is the standard normal variate, $p$ is the estimated proportion to have a particular characteristic and $e$ is the marginal errors of precision. The value of $z$ is depended upon the confidence levels. For example, 95\% confidence level, the value of $z$ is 1.96 while for 99\% the value is 2.58. If a given population is finite then the required sample size can be calculated as,
\begin{equation}
\bar{n}=\frac{n}{1+(n-1)/N}
\end{equation}
where $N$ represents the population size.

\subsection{Optimum allocation}
\begin{figure} [htb]
\centering
\includegraphics[width=1\linewidth]{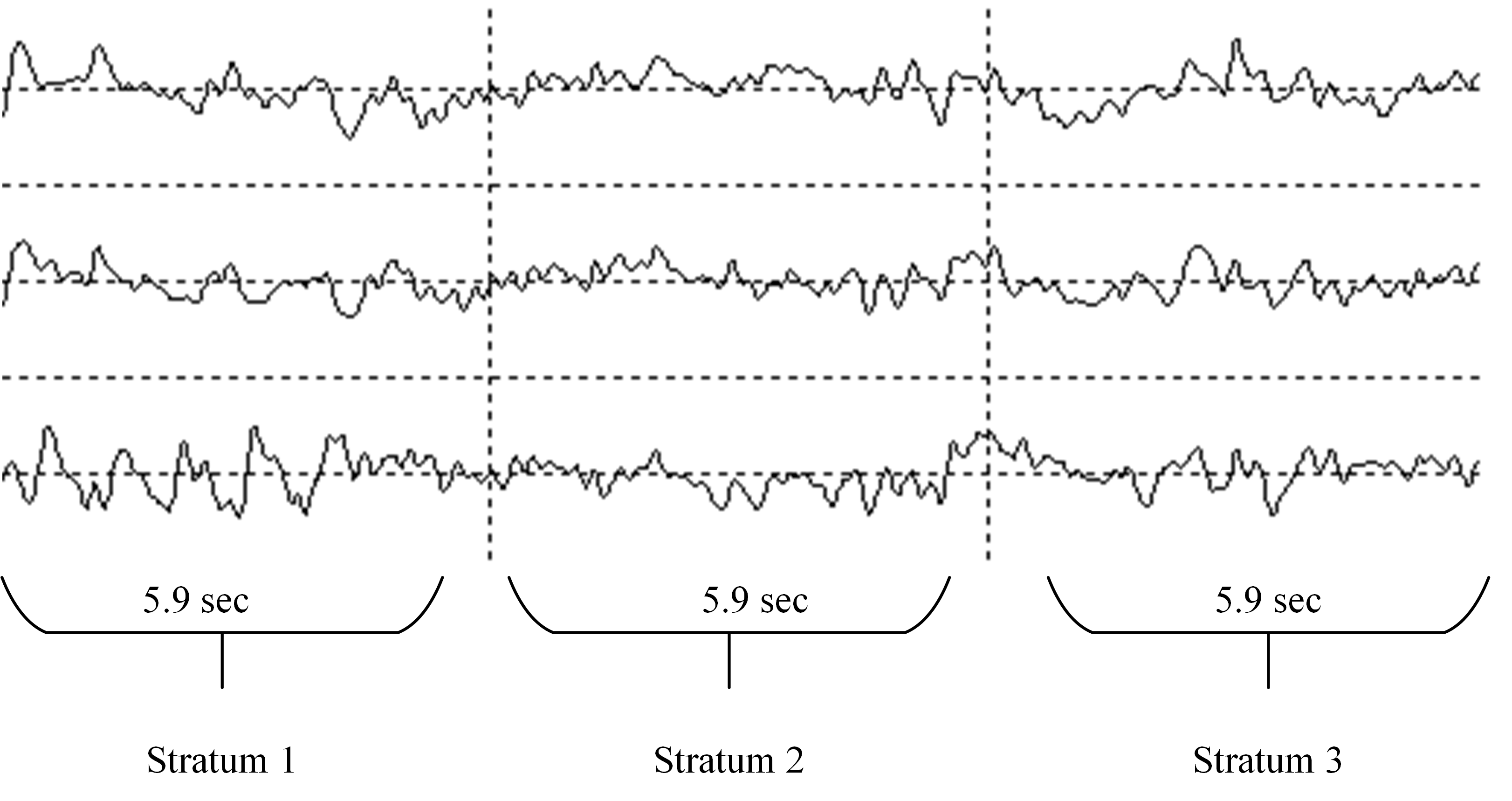}
\caption{\label{f3}The stratification process where the length of each stratum is 5.9 second.}
\end{figure}

Optimum allocation (OA) is a procedure for allocating sample size for each stratum in the population. The allocation process is called optimum because it provides the least variance for estimating a population. After calculating the sample size, we divide the EEG data into different segments or groups named strata. This dividing process is known as stratification. Figure~\ref{f3} shows the stratification process on how the signal is partitioned into strata. The objective of the stratification is to manage the non-stationary properties of the EEG signal. The statistical properties of a non-stationary signal fluctuate over time. However, small windows or segments of those signals show stationarity. Studies show that selecting samples from stratum often increases the accuracy \cite{cochran2007sampling}. For estimating the sample size for each stratum, we applied the OA technique using Equation (3). More details about OA technique is described in references \cite{siuly2015discriminating, li2014novel}.  
\begin{equation}
n_i= \frac{N_i\sqrt{\sum_{j=1}^h s_{ij}^2}}{\sum_{i=1}^k (N_i\sqrt{\sum_{j=1}^h s_{ij}^2})}\bar{n}
\end{equation}
where $n_i$ is the estimated sample size for $i$th stratum, $N_i$ is the data size for $i$th stratum, $s_{ij}^2$ is the variance of the $j$th channel of the $i$th stratum, and $\bar{n}$ is the sample size of the population.
After estimating the sample size for each stratum, we select samples named optimum allocated samples. We then merge each stratum and extract different features from the optimum allocated sample data.

\subsection{Feature extraction}
In this work, we extract different types of features including linear and nonlinear features. The extracted features are minimum, maximum, skewness, mean,  standard deviation, mode, interquartile range, first quartile, third quartile,  Shannon entropy, Hurst exponent, fluctuation index, sample entropy, median, and kurtosis.

\subsection{Feature selection}
Feature selection is a method where the best subset of features is searched automatically from the dataset \cite{peng2016deep}. The search space contains all feasible combinations of features that can be chosen from the given dataset. The goal is to go through the search space and find out the best combination of the feature set that leverage the performance compare to all features. The key benefits of feature selection are threefold: it reduces the overfitting, improves the accuracy, and minimizes the training time. The proposed feature selection algorithm is described in the following paragraph. 

\begin{figure} [htb]
\centering
\includegraphics[width=1\linewidth]{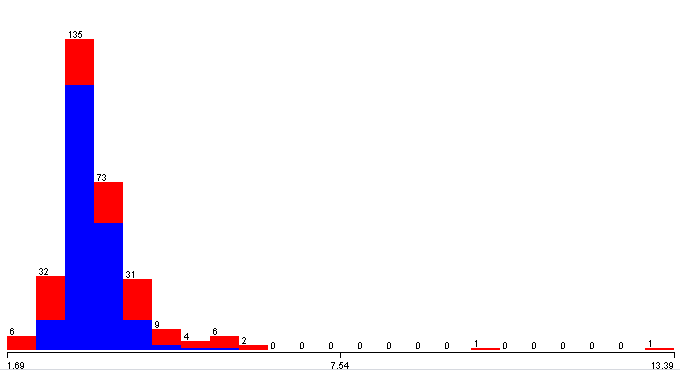}
\caption{\label{f10}Ineffective/weak features for classification (color red represents category 1 and blue represents category 2)(best see in color). }
\end{figure}

\begin{figure} [htb]
\centering
\includegraphics[width=1\linewidth]{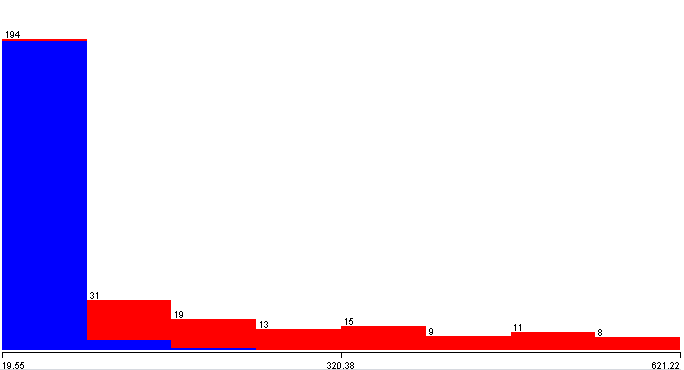}
\caption{\label{f11} Essential feature for classification (color red represents category 1 and blue represents category 2)(best see in color). }
\end{figure}

Research shows that a high-grade feature subsets are highly correlated with the class and uncorrelated with each other. So, to see the feature to feature and feature to class correlation, we use Pearson correlation coefficient \cite{Rodriguez-Lujan:2010:QPF:1756006.1859900},  which can be formally presented as,
\begin{equation}
R_{ij}=\frac{cov(r_i,r_j)}{\sqrt{\sigma^2(r_i).\sigma^2(r_j)}}
\end{equation}
where $r_i$ and $r_j$ are two random variables, $cov$ is the covariance and $\sigma^2$ is the variance. The sample correlation can be calculated using following equation,
\begin{equation}
\bar{R_{ij}}=\frac{\sum_{m=1}^N(r_{mi}-\bar{r_i})(r_{mj}-\bar{r_j})}{\sqrt{\sum_{m=1}^N(r_{mi}-\bar{r_i})^2 \sum_{m=1}^N(r_{mj}-\bar{r_j})^2}}
\end{equation}
  
where $N$ represents the number of samples, $r_{mi}$ shows the $m$-th sample of $r_i$, $\bar{r_i}$ represents the average value of $r_i$. Using the Equation (5), we calculate a matrix that includes correlations between feature to class, and correlations between feature to feature from the all extracted features. We then rank the correlations in decrementing order. Next, we apply the best search technique to get a feature subset with maximum evaluation. For this step, first, we start with a single feature and expand the feature set by adding the next feature. If no enhancement found in the expanded subset outcome, we backtrack and start from the following best-unexpanded subset. If there is no improvement found in the five consecutive expansion, we terminate the search and receive the best feature set. From this feature set, we calculate the mean value of each feature and compute the range value according to the following equation, 
\begin{equation}
range= [\frac{max+min}{2} \pm \frac{max+min}{4}]
\end{equation}
where $min$  and $max$ are the minimum and maximum value of each feature. If 80\% or more data points not present in this range, we eliminate those features from the best feature set. The example of the weak feature and essential feature are shown in Figure~\ref{f10} and Figure~\ref{f11} respectively. After eliminating all the weak features, we get the final feature set for classification. The pseudo code of our proposed feature selection algorithm is shown in Algorithm 1. The novelty of our algorithm is the inclusion of range value. The range value is not calculated in the conventional correlation feature selection algorithm (CFS). More details about CFS is provided in reference \cite{mursalin2017automated}. 

\begin{algorithm}
\caption{Algorithm for Feature Selection}
\begin{flushleft}
\begin{algorithmic} 
\REQUIRE  \emph{Extracted features}
\ENSURE \emph{Best feature sub-set }

\STATE Compute a matrix that includes feature-feature and feature-class correlations from all the extracted features using Equation (5) and rank them in descending order.

\STATE Let, $B$ list contains the $start$ sate, $E$ list is empty, and $T\gets{start}$  

\WHILE {$B$ is not empty }

    \STATE  i. Let, $s = arg\ max( e(x))$ (get the state from $B$ with the maximum assessment)
    \STATE  ii. Delete $s$ from $B$ and include in $E$
    \IF{$e(s)\ge{(T)}$}
        \STATE $T\gets{s}$
    \ENDIF
    \STATE iii. For each leaf node $t$ of $s$ that is not in the $B$ or $E$ list, assess and include in $B$
    \IF{ any change occurs in $T$ during the last set of expansions,}
        \STATE  go to step i 
    \ENDIF
    \STATE iv. $return \ T$
\ENDWHILE
\STATE Calculate the range using Equation 6.

\FOR {$i= {1}\ {to}\  N$ (where $N$ represents the number of features from T list)}
    \IF {80\% or more data points not present in this range,}
        \STATE remove this feature from the feature set
    \ENDIF
\ENDFOR
\STATE $return$ final feature set. 

           \end{algorithmic}
           \end{flushleft}
\end{algorithm}

\subsection{Classification}
In this work, we analyze five different machine learning algorithms for classification including random forest, naïve Bayes, support vector machine, k-nearest neighbor and logistic model trees. The following subsections briefly describe these algorithms.

\subsubsection{Random Forest (RF) classifier}
The RF classifier combines randomized node optimization and bagging \cite{breiman2001random}. It is built by a collection of simple trees called forest which is able to generate a response with a set of predictor values. For standard trees, each node is divided by the best split from all variables. On the other hand, for random forest trees, each node is divided by the best subset from the predictors. RF classifier shows promising performance compared with other well-known classifiers such as discriminant analysis, SVM, neural networks and so on \cite{breiman2001random}.

\subsubsection{Support Vector Machine (SVM)}
SVM is one of the most commonly used machine learning tools that use decision planes to construct decision boundaries \cite{nicolaou2012detection}. A decision boundary divides a set of objects that belong to different classes \cite{ubeyli2010least}. In SVM, the input data is transformed into higher dimensional space followed by the constructed optimal separating hyperplane between different classes. The data vectors closest to the constructed line are known as support vectors. In this paper, we use LIBSVM version 3.21 for classification \cite{chang2011libsvm}. The polynomial kernel function shows better results over different kernels in our experiment. 

\subsubsection{Naive Bayes (NB)}
The core of NB classifier is the Bayesian theorem that is specifically fitted for the high input dimension. This classifier is a simple but effective classifier which can usually outperform more advanced classification methods \cite{mursalin2014towards}. It works by assigning a new observation to the most likely class and considers that the features are conditionally independent with the class value. In a default configuration, Laplace correction is used to prevent the high encounters of zero probabilities \cite{rish2001empirical}\cite{zhang2004optimality}.

\subsubsection{k-Nearest Neighbor ($k$-NN)}
The $k$-NN algorithm uses the similarity measure to classifies new cases from all stored available cases. There are two steps in this algorithm \cite{chaovalitwongse2007time}. 
In the first step, it finds $k$ training samples that are nearest to the invisible sample. In the final step, it takes the commonly occurring classification for these $k$ samples. Then in the regression, it finds the average value of its $k$-nearest neighbors.  The Euclidean distance is used to measure the nearest neighbors. For two given points $Y1=(y11,y12,,,…,y1n)$ and $Y2= (y21,y22,,,…,y2n)$ \cite{song2007iknn}
\begin{equation}
 dist(Y_1,Y_2)=\sqrt{\sum_{i=1}^{N}(y_{1i}-y_{2i})^2}
 \end{equation}

\subsubsection{Logistic Model Trees (LMT)}
The LMT is a classification model based on logistic regression functions. The basic idea of LMT is combining the logistic regression (LR) and decision tree learning. The ordinary decision trees having constants at their leaves can create a piecewise constant model. This algorithm is robust and can handle binary or multi-class target variables. In each node of the tree, the LogiBoost algorithm creates an LR model. This node is divided using the C4.5 criterion \cite{friedman2000additive}\cite{quinlan1993c4}.The pruning technique is used to simplify the model when the tree is expanded completely  \cite{landwehr2005logistic}.   

\subsection{Evaluation}
A ten-fold cross-validation method is used for performance evaluation. This cross-validation reduces the bias of training and test data. To determine the consistency of the experimental results, each and every experiment is repeated twenty times, and the average and standard deviation value are reported. The average accuracy (AC) is calculated using the following equation,  
\begin{equation}
AC=\frac{\sum_{i=1}^{n}w_i x_i}{\sum_{i=1}^{n}w_i}
\end{equation}
where $x$ represents a set of values, and $w$ represents the weight of each data.

\section{Results and discussion }
In this work, we divided the datasets into three categories called healthy (H), interictal (I), and seizure (S). The category H consists of set A and B, category I consist of set C and D, category S consists of set E. We used three different cases for classification: healthy vs. seizure, interictal vs. seizure, and healthy and interictal vs. seizure (see Table~\ref{t1}). 

\begin{table}
\begin{center}
 \caption{\label{t1} Different cases for classification.}
 \begin{tabular}{lllll}
 \toprule
 Case    & Category 1& Category 2& \multicolumn{1}{m{2cm}}{Number of data channels}\\
 \midrule
1&    H    & S & 300\\
2&    I    & S & 300\\
3&    HI    & S & 500\\
 \bottomrule
 \end{tabular}
\end{center}
\end{table}

The sample size for the EEG data was calculated using the Equation (1) and (2). The required sample size using various confidence levels is shown in Table~\ref{sample-size}. Here, the population size $N$ is 4097, $p$ is 0.50, confidence interval is 99-100\%, and $e$ is 0.01. We can reduce 20\% data points using 99\% confidence level while 60\% data points can be reduced using a 70\% confidence level. 

\begin{table}
\begin{center}
 \caption{\label{sample-size} An example of the required sample size for different confidence levels of 4097 data points.}
 \begin{tabular}{lllll}
 \toprule
 Confidence level(\%)    & z value& Sample size & Data reduction (\%)\\
 \midrule
70&    1.04    &1629  &60\\
85&    1.44    &2288  &45\\
95& 1.96    &2872  &30\\
99& 2.58    &3287  &20\\
 \bottomrule
 \end{tabular}
\end{center}
\end{table}

We divided each EEG signal into four strata denoted as Stratum 1 to 4. Each signal contains 4097 data points, so, the first three strata contain 1024 data points, and the last stratum contains 1025 data points. The time duration of each stratum is 5.9 second. The required sample size for each of this stratum was calculated using Equation (3). For example, the sample size for each stratum using 95\% confidence level is shown in Table~\ref{t3}.

\begin{table}
\begin{center}
 \caption{\label{t3} Sample size for each stratum using optimum allocation technique with 95\% confidence level}
 \begin{tabular}{lllll}
 \toprule
 Class    & Stratum 1& Stratum 2& Stratum 3& Stratum 4\\
 \midrule
H (set A)&    696    & 718 & 731 & 727\\
H (set B)&    712&    734& 703& 723\\
I (set C)&    733&    735& 681& 723\\
I (set D)&    724&    727& 688& 733\\
S (set E)& 728    & 737    & 712& 695\\
 \bottomrule
 \end{tabular}
\end{center}
\end{table}

After calculating the number of samples for each stratum, we selected representative samples which are called optimum allocated sample. Next, we extracted 15 different features (see section 2.4) from each optimum allocated stratum. So, the total number of extracted features for each signal is 60. We then applied our proposed feature selection algorithm (see Algorithm 1) for selecting the least number of features. 

\begin{figure}[htb]
\pgfplotstableread[row sep=\\,col sep=&]{
    interval & carT & carD\\
    Case 1     & 98.66  & 98.60\\
    Case 2    & 96.26 & 96.20\\
    Case 3   & 97.08 & 96.96\\
    }\mydata
\begin{tikzpicture}
\tikzstyle{every node}=[font=\small]
    \begin{axis}[
            ybar,
            bar width=.7cm,
            x=2cm, 
            enlarge x limits={abs=1cm}, 
            legend style={at={(0.5,1)},
                anchor=north,legend columns=-1},
            symbolic x coords={Case 1,Case 2, Case 3},
            xtick=data,
            nodes near coords,
            nodes near coords align={vertical},
            ymin=90,ymax=105,
            ylabel={Acuracy (\%)},
        ]
        \addplot table[x=interval,y=carT]{\mydata};
        \addplot table[x=interval,y=carD]{\mydata};
      
        \legend{CFS, Proposed feature selection}
    \end{axis}
\end{tikzpicture}
\caption{\label{f5} Average accuracy comparison between conventional CFS algorithm and our proposed feature selection algorithm using RF classifier with 95\% confidence level.}
\end{figure}
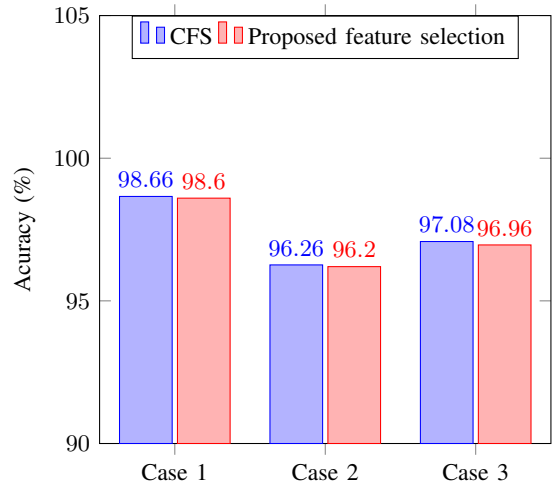

For the performance measurement of our proposed feature selection algorithm, we compared the results with conventional correlation-based feature selection algorithm (CFS) (see Figure~\ref{f4} and Figure~\ref{f5}). We observed that our proposed feature selection algorithm showed similar accuracy with less number of features.

\begin{figure}[htb]
\pgfplotstableread[row sep=\\,col sep=&]{
    interval & carT & carD\\
    Case 1     & 8  & 3\\
    Case 2    & 10 & 7\\
    Case 3   & 7 & 5\\
    }\mydata
\begin{tikzpicture}
\tikzstyle{every node}=[font=\small]
    \begin{axis}[
            ybar,
            bar width=.7cm,
            x=2cm, 
            enlarge x limits={abs=1cm}, 
            legend style={at={(0.5,1)},
                anchor=north,legend columns=-1},
            symbolic x coords={Case 1,Case 2, Case 3},
            xtick=data,
            nodes near coords,
            nodes near coords align={vertical},
            ymin=0,ymax=12,
            ylabel={Number of features},
        ]
        \addplot table[x=interval,y=carT]{\mydata};
        \addplot table[x=interval,y=carD]{\mydata};
      
        \legend{CFS, Proposed feature selection}
    \end{axis}
\end{tikzpicture}
\caption{\label{f4} Comparison between conventional CFS and our proposed feature selection algorithm. Number of required features using RF classifier with 95\% confidence level.}
\end{figure}
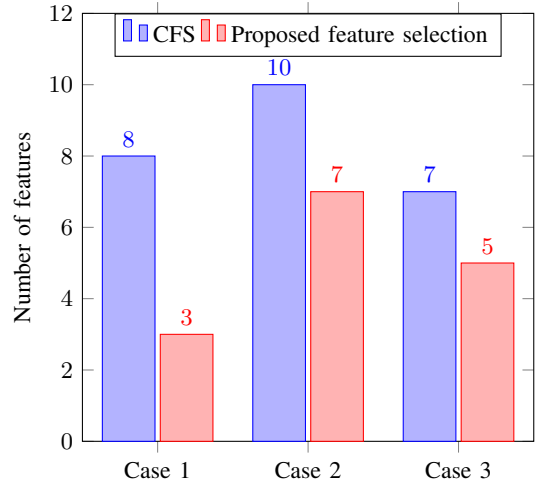

After selecting the required features, we applied five different classifiers. According to our experimental results, RF classifier showed better accuracy compared with other classifiers (see Table~\ref{t4}). The average accuracy comparison of different confidence levels using RF classifier is shown in Table~\ref{t6}.

\begin{table*}[htb]
\begin{center}
 \caption{\label{t4}Average accuracy comparison of different confidence levels using five different classifiers for case 1}
 \begin{tabular}{lllllllll}
 \toprule

\multicolumn{1}{m{1.3cm}}{Confidence levels (\%)}     & \textbf{RF} $\pm$ std& LMT $\pm$ std& k-NN $\pm$ std & SVM $\pm$ std & NB $\pm$ std \\

 \midrule
99 &        98.73 $\pm$ 0.28    &97.53 $\pm$ 0.80 & 91.73 $\pm$ 0.98  & 89.00 $\pm$ 1.79 & 96.73 $\pm$ 0.28\\

95&    98.60 $\pm$ 0.36    &98.53 $\pm$ 0.18 & 95.06 $\pm$ 1.14  & 86.73 $\pm$ 5.01 & 96.00 $\pm$ 1.13\\

85 &    98.46 $\pm$ 0.38    &98.46 $\pm$ 0.18 & 95.86 $\pm$ 1.30  & 89.80 $\pm$ 5.48 & 96.40 $\pm$ 1.48\\

70&    98.39 $\pm$ 0.15    &98.06 $\pm$ 0.49 & 95.27 $\pm$ 1.91  & 90.26 $\pm$ 5.49 & 96.86 $\pm$ 1.24\\

 \bottomrule
 \end{tabular}
\end{center}
\end{table*}

\begin{table*}[htb]
\begin{center}
 \caption{\label{t6}Average accuracy comparison of different confidence levels using RF classifier}
 \begin{tabular}{lllllllll}
 \toprule

\multicolumn{1}{m{1.3cm}}{Confidence levels (\%)}     & Case 1$\pm$ std& Case 2$\pm$ std& Case 3$\pm$ std& \multicolumn{1}{m{1.3cm}}{AC (\%)} \\

 \midrule
99 &        98.73 $\pm$ 0.28    &96.20 $\pm$ 0.50 & 97.4 $\pm$ 0.37  & \textbf{97.44} \\

95&    98.60 $\pm$ 0.36   &96.20 $\pm$ 0.65 & 96.96 $\pm$ 0.38 & 97.20 \\

85 &    98.46 $\pm$ 0.38  &96.00 $\pm$ 0.53 & 96.92 $\pm$ 0.18 & 97.09  \\

70&    98.39 $\pm$ 0.15  &95.86 $\pm$ 0.83 & 96.64 $\pm$ 0.38 & 96.91 \\

 \bottomrule
 \end{tabular}
\end{center}
\end{table*}

\begin{figure}[htb]
\pgfplotstableread[row sep=\\,col sep=&]{
    interval & carT \\
    99\%     & 5.825 \\
    95\%     & 5 \\
    85\%    & 5.1 \\
    70\%    & 6.27 \\
    }\mydata
\begin{tikzpicture}
\tikzstyle{every node}=[font=\small]
    \begin{axis}[
            ybar,
            bar width=.7cm,
            x=1.5cm, 
            enlarge x limits={abs=1cm}, 
            legend style={at={(0.5,1)},
                anchor=north,legend columns=-1},
            symbolic x coords={99\%,95\%, 85\%,70\%},
            xtick=data,
            nodes near coords,
            nodes near coords align={vertical},
            ymin=4,ymax=8,
            ylabel={Number of features},
            xlabel={Confidence levels},
        ]
        \addplot table[x=interval,y=carT]{\mydata};
        
    \end{axis}
\end{tikzpicture}
\caption{\label{f6}Weighted average of required features for different confidence levels using RF classifiers.}
\end{figure}
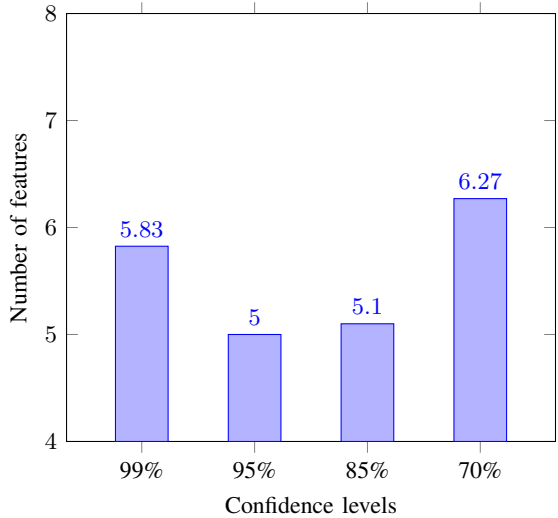

We analyzed the results of different confidence levels and observed that the 95\% confidence level showed the optimum performance. On average only five features required using 95\% confidence level (see Figure~\ref{f6}). Average accuracy comparison of different classifiers using all extracted features and after feature selection with 95\% confidence level is shown in Table~\ref{t5}.

\begin{table*}[htb]
\begin{center}
 \caption{\label{t5}Average accuracy comparison of different classifiers using all extracted features (A)  vs after feature selection (B) with 95\% confidence level}
 \begin{tabular}{lllllllll}
 \toprule
    & & A & &  & & B& &\\
 \midrule
Classifier    & Case 1$\pm$ std& Case 2$\pm$ std& Case 3$\pm$ std& \multicolumn{1}{m{1.3cm}}{AC (\%)} & Case 1$\pm$ std& Case 2$\pm$ std& Case 3$\pm$ std& \multicolumn{1}{m{1.3cm}}{AC (\%)}\\

 \midrule
RF (\%)&        98.40 $\pm$ 0.04    &95.87 $\pm$ 0.18 & 96.96 $\pm$ 0.08  & \textbf{97.06} &98.60 $\pm$ 0.36    &96.20 $\pm$ 0.65 & 96.96 $\pm$ 0.38 &\textbf{97.20}\\

LMT (\%)&    97.33 $\pm$ 0.91   &95.40 $\pm$ 0.59 & 97.16 $\pm$ 0.41  &96.72 &98.53 $\pm$ 0.19    &95.00 $\pm$ 0.78 &96.48 $\pm$ 0.46& 96.74\\

k-NN (\%)&    94.87 $\pm$ 0.55  &92.73 $\pm$ 0.28 & 94.40 $\pm$ 0.79   &94.07 &95.06 $\pm$ 1.14    &94.33 $\pm$ 0.71 &95.56 $\pm$ 0.47& 95.09\\

SVM (\%)&    97.27 $\pm$ 0.43  &94.60 $\pm$ 0.55 & 95.88 $\pm$ 0.56  &95.91 &86.73 $\pm$ 5.01    &92.53 $\pm$ 0.38 &94.60 $\pm$ 0.37 &91.87 \\

NB (\%)&     97.80 $\pm$ 0.18    &92.60 $\pm$ 0.28 & 95.60 $\pm$ 0.14 &95.38 &96.00 $\pm$ 1.13    &92.59 $\pm$ 0.55 &95.72 $\pm$ 0.30 & 94.94\\
 \bottomrule
 \end{tabular}
\end{center}
\end{table*}

\begin{figure} [htb]
\centering
\includegraphics[width=1\linewidth]{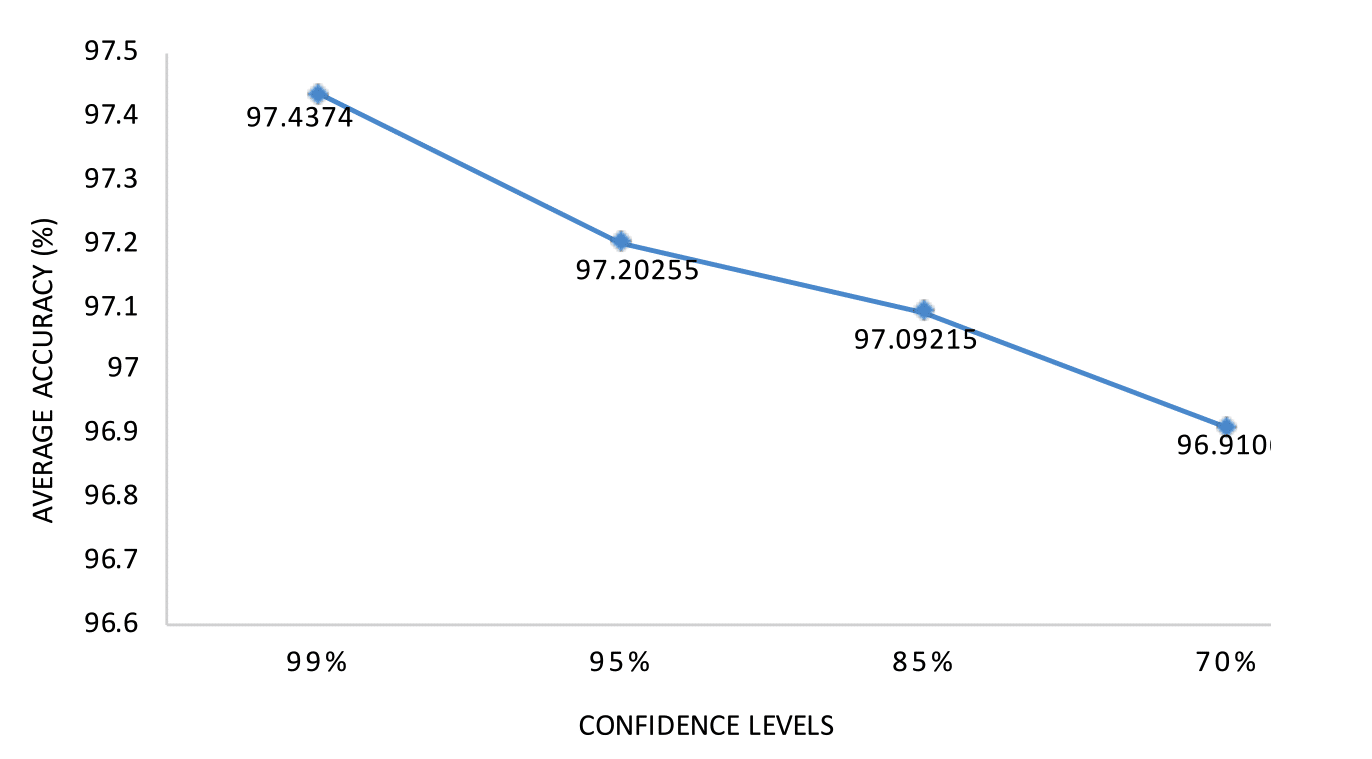}
\caption{\label{f8}Average accuracy comparison using different confidence levels.}
\end{figure}

The point to be noted here is that the accuracy is decreased if we reduce the number of data points. Our experimental results suggested that we should not reduce the data points beyond 30\% which corresponds to a confidence level of 95\% (see Figure~\ref{f8}).  

Recently, Acharya et al. \cite{acharya2018deep} proposed a deep convolutional neural network-based method to detect seizure from the EEG signal. However, they reported a poor classification accuracy (88.7\%) compared with the other state-of-the-art methods. They need more training data to improve performance while our method shows better results with reduced data points. Zahra et al. \cite{zahra2017seizure} presented a method using multivariate EMD and artificial neural network, which is computationally expensive and gained an overall classification accuracy of 87.2\%. The performance of some state-of-the-art methods is reported in Table~\ref{t7}. We select only those methods that use the same database \cite{andrzejak2001indications} and the same cases for epileptic seizure classification. Our proposed method demonstrates comparable performance even after 30\% reduction in data points.

\begin{table*}[htb]
\begin{center}
 \caption{\label{t7}Contrast among different state-of-the-art methods and our proposed method.}
 \begin{tabular}{llll}
 \toprule
 Studies&    Method&    Cases & Accuracy(\%)\\
 \midrule
Sahbi et al. \cite{CHAIBI2013927}&Hilbert huang transform and rms features&Case 1& 90.72\\


Yatindra et al. \cite{kumar2014epileptic} & Fuzzy approximate entropy and SVM & Case 3& 97.38\\

Varun et al. \cite{joshi2014classification}& Fractional linear prediction and SVM& Case 2& 95.33\\

Maheshkumar et al.  \cite{kolekar2015nonlinear} & Non-linear feature  using least square support vector machine & Case 1& 91.25\\
& & Case 2& 83.75\\

This work& RF classifier with proposed feature selection (using 95\% confidence level) & Case 1& 98.60\\
& & Case 2 & 96.20\\
& & Case 3& 96.96\\

 \bottomrule
 \end{tabular}
\end{center}
\end{table*}

\section{Conclusion}
In this work, we present an efficient, cost-effective method for epileptic seizure classification. The objective is to identify the effectiveness of the data reduction using the representative sample data based on the sampling technique. In order to reduce the computational cost, we have also proposed a feature selection algorithm based on correlation and the threshold, which provides better performance compared with the conventional correlation-based feature selection algorithm. Finally, five different machine learning algorithms are used for the classification of the selected features. The experimental results show that our proposed method using sampling technique and feature selection algorithm along with the Random Forest Classifier can be an effective solution for epileptic seizure classification.  
\section*{Acknowledgments}
This work is supported by the ECU Higher Degree by Research Scholarship, ECU-ECR, and Strategic Initiative Fund grants. The authors would like to thank Professor David Suter for his valuable feedback. 
\bibliographystyle{IEEEtran}
\bibliography{report}
  \begin{wrapfigure}{l}{15mm} 
    \includegraphics[width=1in,height=1in,clip,keepaspectratio]{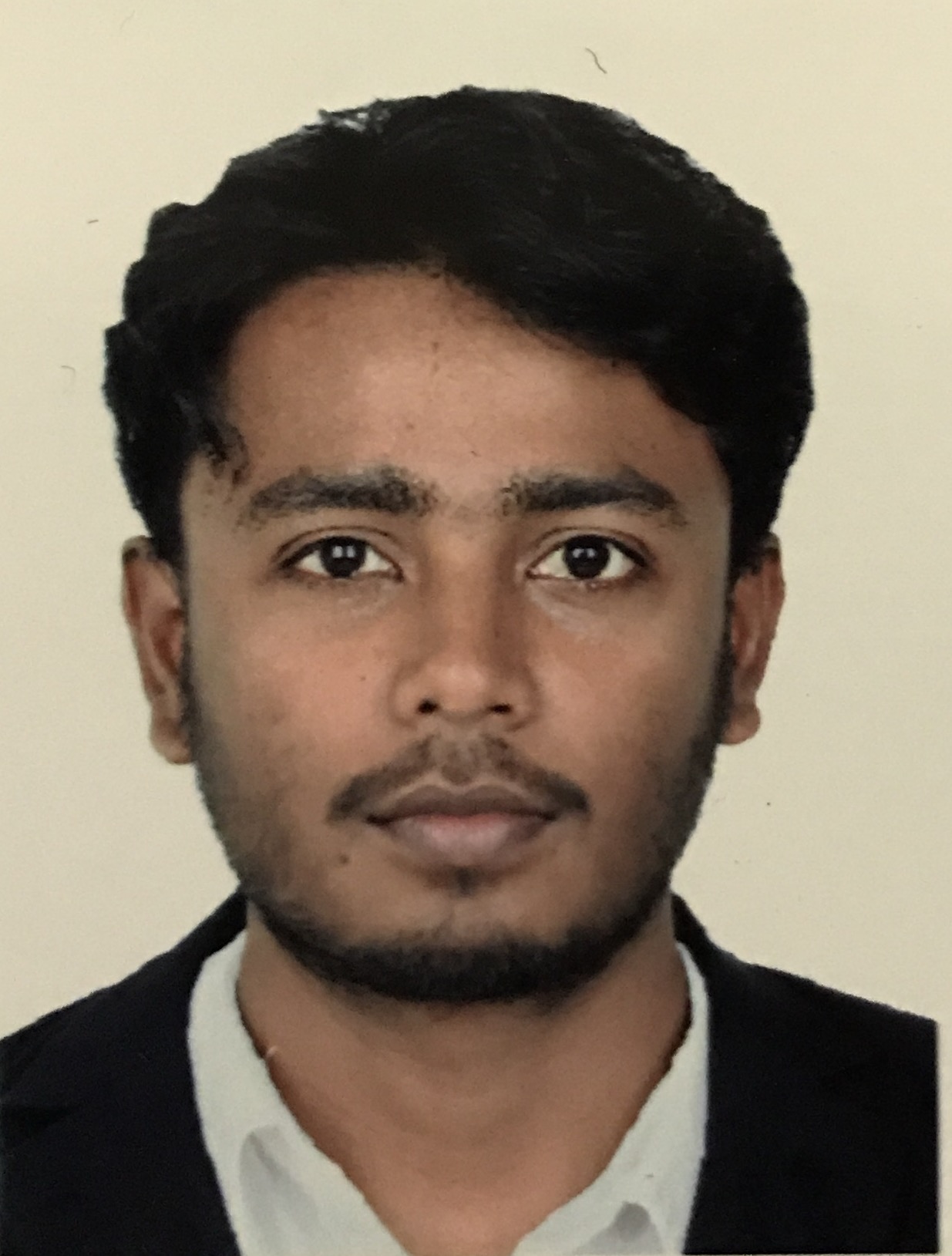}
  \end{wrapfigure}\par
\textbf{Md Mursalin} is currently a research fellow at Edith Cowan University. He received his M.Sc. in computer science and engineering from the University of Jinan, China, and B.Sc. in Computer Science and Information Technology from Islamic University of Technology (IUT), Bangladesh. He worked as an Assistant Professor in Computer Science at Pabna University of Science and Technology, Bangladesh. He has several publications in different reputed journals and conferences. His research interests span across biomedical signal processing, brain-computer interface, human-computer interaction.  \par
    
  \begin{wrapfigure}{l}{15mm} 
    \includegraphics[width=1in,height=1in,clip,keepaspectratio]{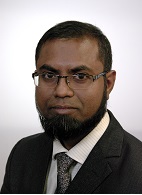}
  \end{wrapfigure}\par
\textbf{Dr Islam} completed his PhD with Distinction in Computer Engineering from the University of Western Australia (UWA) in 2011. He received his MSc in Computer Engineering from King Fahd University of Petroleum and Minerals in 2005 and BSc in Electrical and Electronic Engineering from Islamic Institute of Technology in 2000. He was a Research Assistant Professor at UWA from 2011 to 2015, a Research Fellow at Curtin University from 2013 to 2015 and a Lecturer at UWA from 2015-2016. Since 2016, he has been working as Lecturer in Computer Science at Edith Cowan University. He has published around 52 research articles and got nine public media releases. He obtained 14 competitive research grants for his research in the area of Image Processing, Computer Vision and Medical Imaging. He has co-supervised to completion seven honours and postgrad students and currently supervising one MS and six PhD students. He is serving as Associate Editor of 13 international journals, Technical Committee Member of 24 conferences and regular reviewer of 26 journals. He is also serving seven professional bodies including IEEE and Australian Computer Society (Senior Member).
\par
\begin{wrapfigure}{l}{15mm} 
    \includegraphics[width=.8in,height=.8in,clip,keepaspectratio]{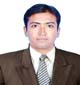}
  \end{wrapfigure}\par
\textbf{Md Kislu Noman} received B.Sc.and M.Sc. degrees in Computer Science and Engineering from Islamic University, Bangladesh. He is now working as Assistant Professor at the Department of Computer Science and Engineering, Pabna University of Science and Technology, Bangladesh. His research interest areas are Brain Computer Interfacing (BCI), Bioinformatics, Image processing, Signal Processing and Wireless communications.   
  \par
  
  \begin{wrapfigure}{l}{15mm} 
    \includegraphics[width=1in,height=1in,clip,keepaspectratio]{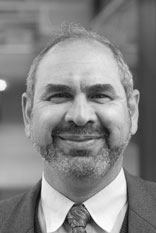}
  \end{wrapfigure}\par
\textbf{Dr Adel Al-Jumaily} is Associate Professor in the University of Technology Sydney. He is holding a Ph.D. in Electrical Engineering (AI); He is working in the cross-disciplinary applied research area and established a strong track record. He established and led many research groups and delivered many projects, in addition to his contributions in building and extending many laboratories in this area. He has published more than 200 peer-reviewed publications, got 7 patents, and supervised to graduation more than 11 higher degree students and 2 Supervision Awards, acquired 7 million AUD, 5 best papers awards, and many high achievements. His research interests include Computational Intelligence, Bio- Mechatronics Systems, Health Technology and Biomedical, Vision based cancer diagnosing, Bio-signal/ image pattern recognition, and Artificial Intelligent Systems.   
  \par
\end{document}